\documentclass[conference]{IEEEtran}
\IEEEoverridecommandlockouts
\usepackage{cite}
\usepackage{amsmath,amssymb,amsfonts}
\usepackage{algorithmic}
\usepackage{graphicx}
\usepackage{textcomp}
\usepackage{xcolor}
\usepackage{url}
\usepackage{balance}
\def\BibTeX{{\rm B\kern-.05em{\sc i\kern-.025em b}\kern-.08em
    T\kern-.1667em\lower.7ex\hbox{E}\kern-.125emX}}
\begin{document}

\title{Lang2Manip: A Tool for LLM-Based Symbolic-to-Geometric Planning for Manipulation
\thanks{$^{1}$ Khalifa University Center for Autonomous Robotic Systems (KUCARS), Khalifa University, United Arab Emirates.}%
 \thanks{$^{2}$ Institute of Industrial and Control Engineering (IOC), Universitat Politècnica de Catalunya, Spain.}
\thanks{$^{*}$ This publication is based upon work supported by the Khalifa University of Science and Technology under Award No. RC1-2018-KUCARS, and Project PID2024-157729OB-I00 funded by MICIU/AEI/10.13039/501100011033/FEDER, UE.\\ 
 Corresponding Author, Email: irfan.hussain@ku.ac.ae}
}

\author{Muhayy Ud Din$^{1}$, Jan Rosell$^{2}$,  Waseem Akram$^{1}$,  and Irfan Hussain$^{1,*}$

}

\maketitle

\begin{abstract}
Simulation is essential for developing robotic manipulation systems, particularly for task and motion planning (TAMP), where symbolic reasoning interfaces with geometric, kinematic, and physics-based execution. Recent advances in Large Language Models (LLMs) enable robots to generate symbolic plans from natural language, yet executing these plans in simulation often requires robot-specific engineering or planner-dependent integration. In this work, we present a unified pipeline that connects an LLM-based symbolic planner with the Kautham motion planning framework to achieve generalizable, robot-agnostic symbolic-to-geometric manipulation. Kautham provides ROS-compatible support for a wide range of industrial manipulators and offers geometric, kinodynamic, physics-driven, and constraint-based motion planning under a single interface. Our system converts language instructions into symbolic actions and computes and executes collision-free trajectories using any of Kautham’s planners without additional coding. The result is a flexible and scalable tool for language-driven TAMP that is generalized across robots, planning modalities, and manipulation tasks.
\end{abstract}

\IEEEpeerreviewmaketitle

\section{Introduction}

Kinamatic and dynamic simulation plays a central role in modern robotics, providing a safe, efficient, and scalable environment for developing algorithms before transferring them to real hardware. Whether for manipulation, locomotion, or multi-robot coordination, simulation environments allow to test perception pipelines, control strategies, and complex interaction behaviors with repeatability and without the risk of hardware damage. As robotic systems become more capable and more diverse, simulation has become a critical foundation for benchmarking algorithms, validating design assumptions, and accelerating prototyping across a broad spectrum of applications \cite{koenig2004gazebo,todorov2012mujoco,coumans2021pybullet}.

Task and Motion Planning (TAMP) is one domain where simulation is particularly important. TAMP requires both high-level reasoning about task structure (e.g., sequencing pick-place, rearrangement, or assembly steps) and low-level geometric reasoning for grasping, inverse kinematics, collision avoidance, and trajectory generation. Therefore, the development and evaluation of TAMP pipelines is dependent on simulation tools that provide accurate geometric models, reliable physics, and flexible robot and environment configurations. As TAMP has progressed from classical symbolic planners to more data-driven and hybrid approaches, simulators have become deeply embedded in the workflow of evaluating planning pipelines and 
studying how high-level symbolic decisions connect to executable motion \cite{erol1994htn,garrett2020pddlstream}.

Existing tools, such as PyBullet~\cite{coumans2021pybullet} and MoveIt~\cite{chitta2012moveit}, 
offer strong capabilities for kinematic modeling, dynamics simulation, and 
collision-aware motion generation. These frameworks are widely used for 
manipulation research and increasingly underpin recent advances in 
language-conditioned task planning, where Large Language Models (LLMs) generate symbolic task descriptions that are executed in simulation. Several studies have extended PyBullet or similar environments for language-guided manipulation, demonstrating zero-shot planning results \cite{wang2024llm3,uddin2024ontologydriven,huang2022inner}. However, such extensions are often tightly coupled to a specific robot model, scene structure, or planner configuration, making it challenging to generalize the pipeline across different manipulators, domains, or motion planning paradigms. When introducing new robots, 
additional coding effort, custom configuration files, or new URDF conversions are typically required. Similarly, the swapping of geometric planners or the incorporation of kinodynamic or constraint-based planning modules usually requires substantial changes to the underlying pipeline \cite{ratliff2009chomp,schulman2014trajopt,sucan2012ompl}.

\begin{figure*}[t]
\centering
\includegraphics[width=\linewidth]{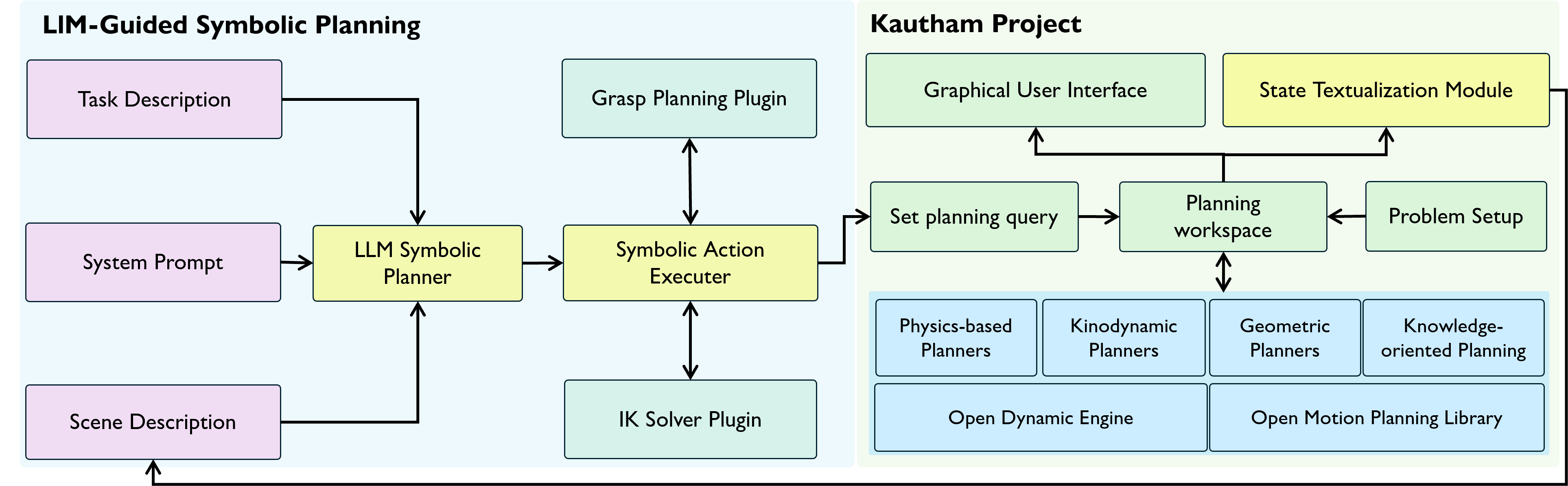}
\caption{Overview of the proposed LLM–Kautham manipulation planning framework. 
The LLM generates symbolic actions from task, system, and scene descriptions, 
which are executed via grasp and IK plugins. These actions are passed to the 
Kautham Project, which handles problem setup and motion planning through 
its GUI, state textualization module, and multiple planners 
(physics-based, kinodynamic, geometric, and knowledge-oriented).}
\label{fig:framework}
\end{figure*}
In this work, we propose a system-level tool that bridges LLM-based symbolic 
planning with the \emph{Kautham} motion planning framework, enabling 
language-conditioned manipulation that is naturally adaptable across robots. 
planning modalities and domains. Kautham~\cite{rosell2014kautham} is a platform that uses sampling-based planners via Open Motion Planning Library (OMPL)~\cite{sucan2012ompl}. It is compatible with ROS, designed to support a variety of industrial manipulators such as KUKA, ABB YuMi, UR5, Franka Emika Panda, and Staubli robots. It offers a standardized interface for planning that includes geometric, kinematic, kinodynamic, physics-driven, and constraint-based approaches.
Using Kautham's rich set of planners and its seamless handling of robot 
and scene models, our system executes LLM-generated symbolic plans without 
requiring task-specific coding, custom integrations, or planner-dependent 
modifications. This allows a single LLM-generated symbolic plan to be executed using multiple planning strategies and robotic platforms, significantly improving the scalability and generality of language-driven manipulation systems.

Our contribution is therefore a generalizable, modular, and robot-agnostic 
pipeline that demonstrates how LLMs can be connected to advanced symbolic to geometric planning capabilities through a mature motion planning framework. The developed system provides a versatile basis for research in LLM-conditioned manipulation, task and motion planning, and robot learning. It allows the utilization of Kautham’s comprehensive planning capabilities without requiring additional engineering efforts.

\section{Related Work}

\subsection{Simulation Frameworks for Manipulation and Planning}
A wide range of kinematic and dynamic simulation tools support robotic manipulation and motion planning,
offering varying levels of realism and integration with
planning libraries. Gazebo \cite{koenig2004gazebo}, MuJoCo \cite{todorov2012mujoco},
PyBullet \cite{coumans2021pybullet} and Isaac Gym \cite{macklin2020isaac} are
commonly used platforms that provide rigid-body dynamics, contact modeling, and
configurable robot and environment representations. These simulators form the
basis for the evaluation of control policies, geometric planners, and learning-based
approaches. However, their use in integrated task-and-motion planning pipelines
typically requires manual configuration of robot models, custom interfaces, and
tool-specific adaptation of planning modules.

\subsection{Task and Motion Planning Frameworks}
Classical TAMP research has focused on unifying symbolic task reasoning with
continuous motion generation, using formalisms such as STRIPS \cite{fikes1971strips},
HTN planning \cite{erol1994htn}, or methods like PDDLStream
\cite{garrett2020pddlstream}. Sampling and optimization based planners,
including RRT \cite{lavalle1998rapidly}, PRM \cite{kavraki1996probabilistic}, CHOMP
\cite{ratliff2009chomp}, and TrajOpt \cite{schulman2014trajopt} have been widely
implemented in MoveIt \cite{chitta2012moveit} and OMPL \cite{sucan2012ompl}, which
serve as standard toolchains for motion generation. Despite their flexibility,
deploying these frameworks for symbolic task-level integration often requires
additional engineering for model integration, scene setup, and planner selection.
The Kautham framework\footnote{\url{https://github.com/iocroblab/}} \cite{rosell2014kautham} provides a ROS-compatible environment with built-in support for various industrial
manipulators, geometric~\cite{Rosell619500415} and kinodynamic planners, physics-based planning~\cite{Gillani201643}, and even extend to
knowledge-oriented planning capabilities \cite{muhayyuddin2018kpmp}.

\subsection{LLM-Based Planning and Language-Driven Manipulation}
Recent advances in Large Language Models have enabled language-conditioned
manipulation and high-level task planning. Approaches such as ProgPrompt \cite{singh2023progprompt} and Inner Monologue
\cite{huang2022inner} use LLMs to generate symbolic plans or hierarchical task
structures. More recent work integrates LLM planning with feasibility reasoning, such as LLM\(^3\)
\cite{wang2024llm3}, or with ontologies for symbolic grounding
\cite{uddin2024ontologydriven}. However, these systems are implemented in
specific simulation tools, such as PyBullet or custom environments, and tend to be
tightly coupled to particular robot models or motion-planning backends.

The proposed system differs from prior work by coupling LLM-generated symbolic
plans with the Kautham framework, enabling access to a varity of planners within a unified, robot-agnostic
infrastructure. Rather than extending a single simulator, our pipeline utilizes
Kautham’s extensive planning capabilities and support for industrial robots, allowing
language-conditioned TAMP to generalize between robots, planning modalities, and
simulation setups with minimal integration effort.

\section{Lang2Manip Framework}

The proposed framework integrates LLM-guided symbolic planning with the Kautham project to produce a unified pipeline capable of executing language-conditioned manipulation tasks across multiple robots and planning paradigms. As illustrated in Fig.~\ref{fig:framework}, the architecture is organized into two main subsystems: the LLM-guided symbolic planning layer and the Kautham-based motion planning and execution layer. Together, these components enable the transformation of high-level natural-language instructions into executable robot trajectories in a structured and robot-agnostic manner.

\subsection{The Kautham Project}
The Kautham Project \cite{rosell2014kautham} is an open-source framework for robotic motion planning that brings together a wide range of planning algorithms and modeling utilities. It is built on top of the Open Motion Planning Library (OMPL), which provides the core sampling-based planners, collision-checking routines, and environment modeling tools. Kautham extends these capabilities with a flexible interface for defining robots, obstacles, and planning scenes. For communication and integration with external components, Kautham includes a ROS interface (the \texttt{kautham\_ros} package), which allows the framework to be embedded into larger robotic systems.

\begin{figure}[t]
\centering
\includegraphics[width=\columnwidth{}{}{}]{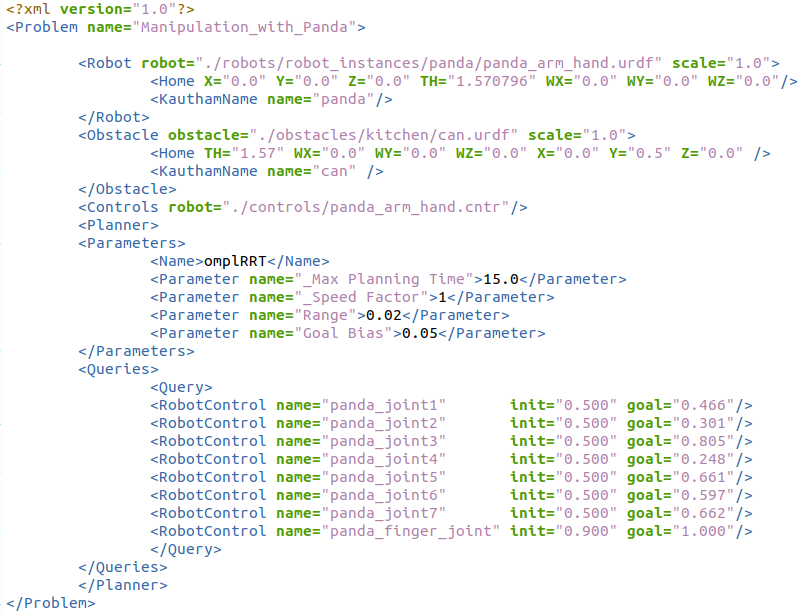}
\caption{Example Kautham problem file defining a manipulation scene with the Franka Emika Panda robot. 
The XML specification includes the robot and obstacle URDF models, their poses in the workspace, 
the associated control file, planner selection (RRT in this example), planner parameters, 
and a query block specifying the initial and goal values for each controlled joint.}

\label{fig:problem-file}
\end{figure}
\subsubsection{Robot and Scene Integration}

The proposed framework relies on the Kautham Project to seamlessly integrate  robot models, scene descriptions, and collision environments into the motion-planning pipeline. Kautham provides native support for URDF import, allowing complete kinematic chains, joint limits, link geometries, and collision models to be loaded directly from standardized robot descriptions. Scene objects are likewise defined as URDF models. The workspace (planning scene) is specified through XML-based problem files, in which each element is referenced by its corresponding URDF model together with its pose in the environment as depicted in Fig.~\ref{fig:problem-file}. During initialization, Kautham constructs a unified configuration space by combining the robot’s joint representation with the poses of  obstacles, ensuring that all components, robots, meshes, manipulatable items, and collision bodies, are consistently registered within the planning workspace.

\begin{figure}[t]
\centering
\includegraphics[width=\columnwidth{}{}{}]{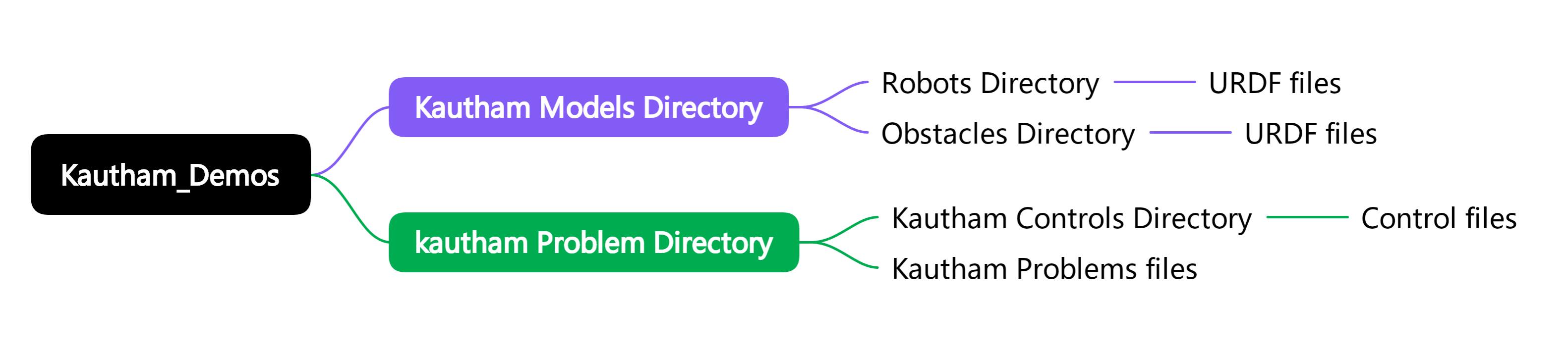}
\caption{Directory structure used in Kautham for organizing planning scenes. 
The \texttt{Kautham\_Demos} folder contains a \texttt{Kautham Models Directory} 
with URDF files for robots and obstacles, and a \texttt{Kautham Problem Directory} 
with problem descriptions and control files required to instantiate a planning scenario.}

\label{fig:dstructure}
\end{figure}

These problem files and URDF models are organized in a  directory structured, as illustrated in Fig.~\ref{fig:dstructure}. Robot and object models are placed in a dedicated model directory, while a corresponding problem directory contains the XML files describing the robot instance, obstacle set, initial and goal configurations, and available control modes. All model references are defined using relative paths, preserving platform independence and supporting scene portability across systems. From these specifications, Kautham automatically generates its internal problem structure and associated control configuration files, enabling the planner to correctly interpret robot capabilities and valid interaction constraints. This structured organization of models and problem files improves reproducibility, simplifies reuse across tasks, and provides a reliable foundation for integrating LLM-generated symbolic actions with the geometric planning backend.

\begin{figure*}[t]
\centering
\includegraphics[width=\linewidth{}{}{}]{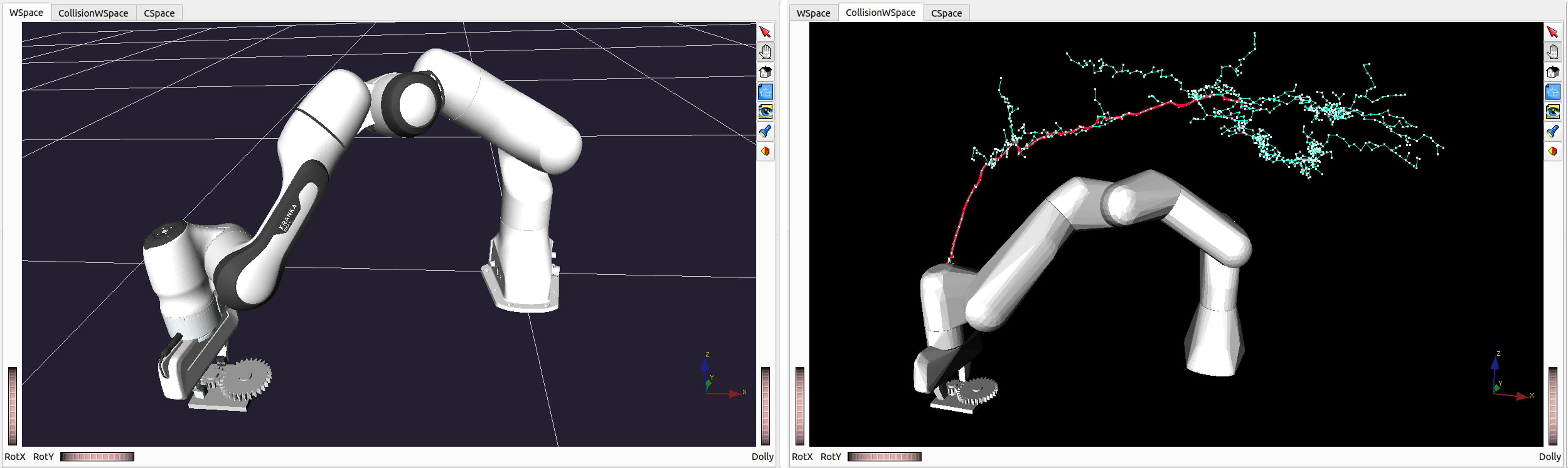}
\caption{Kautham visualization of a manipulation scene. 
The left panel shows the standard Kautham viewer displaying the robot and workspace objects, 
while the right panel illustrates the robot's collision model together with the computed motion plan 
(red trajectory) and the corresponding exploration tree generated by the RRT planner (green samples).
}

\label{fig:gui}
\end{figure*} 

\subsubsection{Graphical User Interface and Execution Flow}

The graphical user interface facilitates the planning workflow by visualizing robot configurations, object states, validated trajectories, and planning outcomes. This facilitates debugging, monitoring, and detailed analysis of both symbolic and geometric planning steps.  The Kautham has two complementary visualization components: A Qt5-based interface and an RViz-based viewer for \texttt{kautham\_ros} package. Together, they provide a detailed visualization of the planning scene, sampled configurations, collision geometries, and executed trajectories. Fig.~\ref{fig:gui} shows example screenshots of the Qt5-based visualization, illustrating how robot models, obstacles, and planned motions are displayed during the planning process.

\subsubsection{Environment State Observation}
\label{sec:textualization}

This feature serves as a critical component for bridging the gap between the geometric workspace representation and symbolic reasoning systems in the \textit{Lang2Manip} framework. This method systematically converts the internal workspace state into a structured textual description that includes comprehensive information about robots, obstacles, and environmental constraints. The function extracts detailed information such as robot joint configurations, link hierarchies, DOF specifications, obstacle positions and orientations, bounding box dimensions, and spatial relationships between objects. By providing both geometric properties (positions, orientations, dimensions) and semantic classifications (robot names, joint names, obstacle types), this textualized state observation enables Large Language Models to understand the current workspace state and generate contextually appropriate task plans. The structured output format ensures that spatial reasoning algorithms can process object relationships, collision constraints, and manipulation feasibility, making it an essential interface between the geometric motion planning domain and symbolic task planning systems.

\subsection{LLM-Guided Symbolic Planning}

\subsubsection{Action Grammar and Symbolic Action Space}

To ensure consistent and machine-interpretable plan generation, the LLM operates in a predefined symbolic action space 
\(\mathcal{A}\). The action grammar constrains the LLM to produce actions drawn from a fixed vocabulary of manipulation 
primitives. In our framework, the symbolic action set is defined as:
\[
\mathcal{A} = \{\texttt{pick},\, \texttt{place},\, \texttt{move},\, \texttt{push}\}.
\]
Each symbolic action \(a \in \mathcal{A}\) follows a structured template with a fixed argument list, 
ensuring that all LLM-generated outputs conform to a consistent grammar. 
We express a generic action in the symbolic space as:
\[
a(o,\, \mathbf{p},\, \mathbf{r},\, \kappa),
\]
where \(o\) denotes the target object of the action, \(\mathbf{p}\) represents optional positional 
or goal parameters (e.g., target pose for placement or target waypoint for movement), 
\(\mathbf{r}\) encodes action-specific refinements such as grasp direction or approach vector, 
and \(\kappa\) specifies the preferred motion planner (e.g., RRT, RRTConnect) to be used during execution. 
Different actions from \(\mathcal{A}\) instantiate this general structure with varying semantic 
requirements, e.g., \texttt{pick} uses \(\mathbf{r}\) to indicate grasp direction, while 
\texttt{place} uses \(\mathbf{p}\) to specify a placement pose.
This unified grammar ensures that all symbolic plans produced by the LLM can be reliably parsed 
and grounded into geometric operations during execution in Kautham.

\begin{figure*}[t]
\centering
\includegraphics[width=\linewidth]{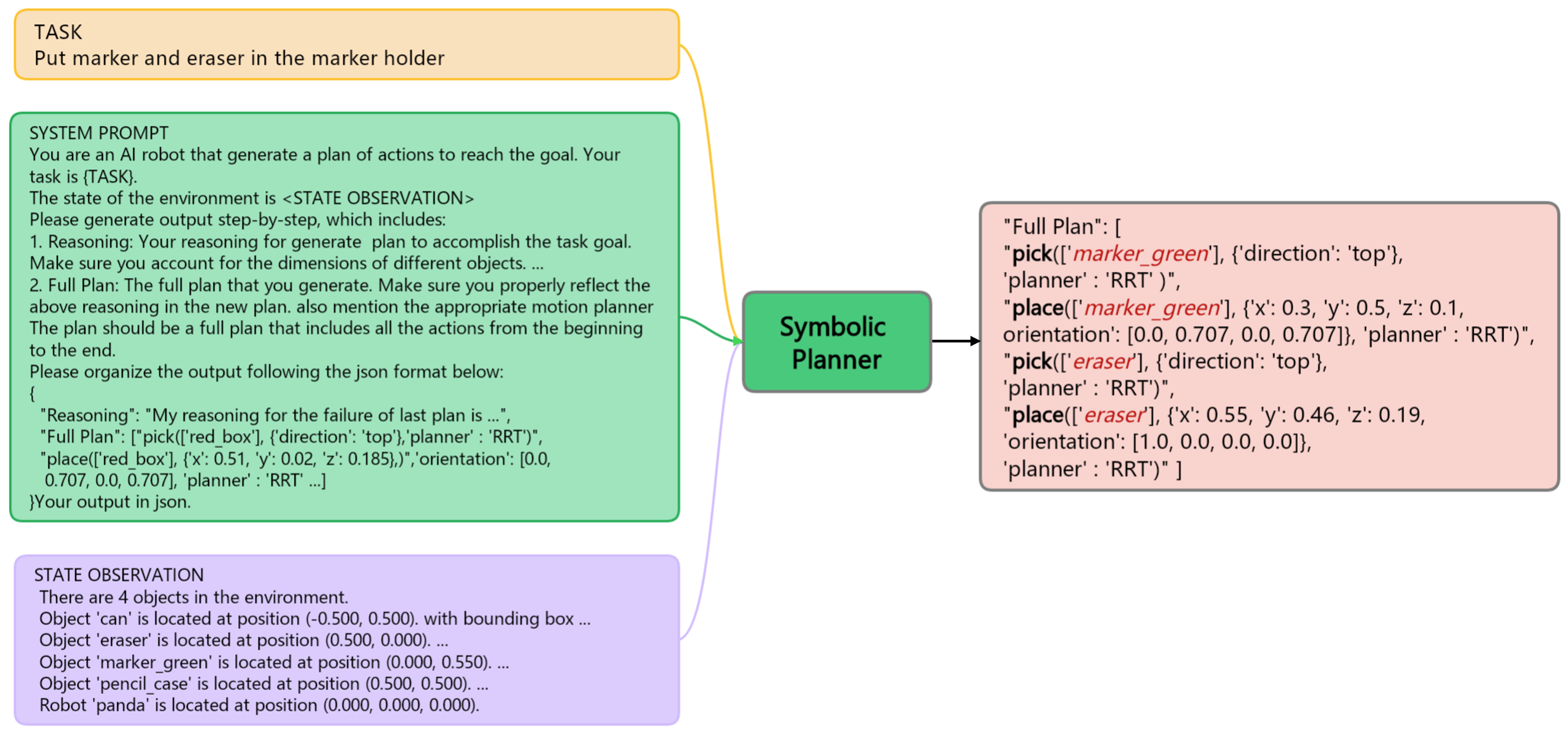}
\caption{LLM-guided symbolic planning pipeline. The prompt is composed of three components: the user-defined task, the fixed system prompt describing the required output format and action schema, and the textualized environment state obtained from Kautham. The combined prompt is passed to the LLM, which produces a structured JSON plan containing high-level symbolic actions.}

\label{fig:structure}
\end{figure*}
\subsubsection{Prompting and symbolic plan Generation}
The symbolic planning layer converts a natural-language task description into a structured sequence of high-level symbolic actions. In our framework, the prompt provided to the LLM is composed of three essential components. First, the \textit{task description}, provided by the user, specifies the manipulation goal in natural language (e.g., ``put the marker and eraser in the holder’’). Second, a fixed \textit{system prompt} defines the overall planning context, the symbolic action schema, and the required JSON output format. This ensures that the LLM consistently generates interpretable symbolic plans. Third, the \textit{state observation} is a textualized description of the current environment obtained from Kautham’s state observation module, listing objects, their poses, and other relevant spatial properties in natural-language form.

These three components are concatenated into a single integrated prompt and passed to the LLM-based symbolic planner ash depicted in Fig.~\ref{fig:structure}. Although GPT-4 is used in our experiments, the framework is model-agnostic and compatible with any modern LLM capable of structured output generation, such as Gemini, Llama, and Claude. The LLM interprets the task and environment, reasons over object arrangements, and outputs a symbolic plan in JSON format. Each plan consists of a sequence of high-level actions (e.g., \texttt{pick}, \texttt{place}) along with their associated parameters such as target object, placement pose, or preferred planner. This plan is robot-independent, focusing solely on the logical sequence of actions required to achieve the task goal rather than geometric feasibility.

The resulting symbolic plan is forwarded to the execution pipeline, where each action is grounded using grasp planning, inverse kinematics, and motion planning. This modular design allows LLMs to serve as flexible task planners without requiring predefined PDDL domains and hand-coded symbolic rules. The only information the LLM requires is the task objective, a reusable system prompt, and the automatically textualized environment state.

\section{Symbolic to Geometric Execution }

The symbolic plan produced by the LLM serves as the starting point for the complete geometric planning and execution pipeline. Each symbolic action generated in JSON format encodes the high-level objective of the task, such as \texttt{pick}, \texttt{place}, \texttt{move}, or \texttt{push}, together with the relevant parameters, including the target object, grasp direction, desired placement pose, and the preferred motion planner. Although these actions provide a structured representation of the task, they remain abstract and independent of robot kinematics, workspace geometry, and collision constraints. The core function of the execution pipeline is, therefore, to transform these
symbolic descriptions into precise motion planning queries to be sent to the Kautham planner.

This transformation begins with the interpretation of each symbolic action. The system parses the JSON action entry, extracts the action type and parameters, and queries the object registry to obtain the precise geometric information needed for planning. This includes the object’s current pose, dimensions, and its spatial relationships with surrounding obstacles. By grounding symbolic references in actual geometric properties, the system ensures that every subsequent computation is physically meaningful and aligned with the real workspace configuration.

\subsection{Grasp Pose Computation}
For actions requiring object acquisition, such as \texttt{pick}, the next stage involves grasp computation. 
The grasp planner analyzes the object’s geometry, the specified grasp direction, and any scene constraints 
to compute a feasible end-effector pose. In principle, the framework is compatible with any state-of-the-art 
grasp planning module, as the grasp planner is implemented as an interchangeable plugin. For the experiments 
presented in this work, we developed a lightweight grasp planner that selects a grasp pose based on the 
approach direction specified in the symbolic plan (e.g., top grasp, side grasp), and applies appropriate 
offsets and safety margins to ensure a collision-free approach. Once a valid grasp pose is computed, the 
inverse kinematics module converts this Cartesian target into a robot joint configuration.

\begin{figure*}[t]
\centering
\includegraphics[width=\textwidth{}{}{}]{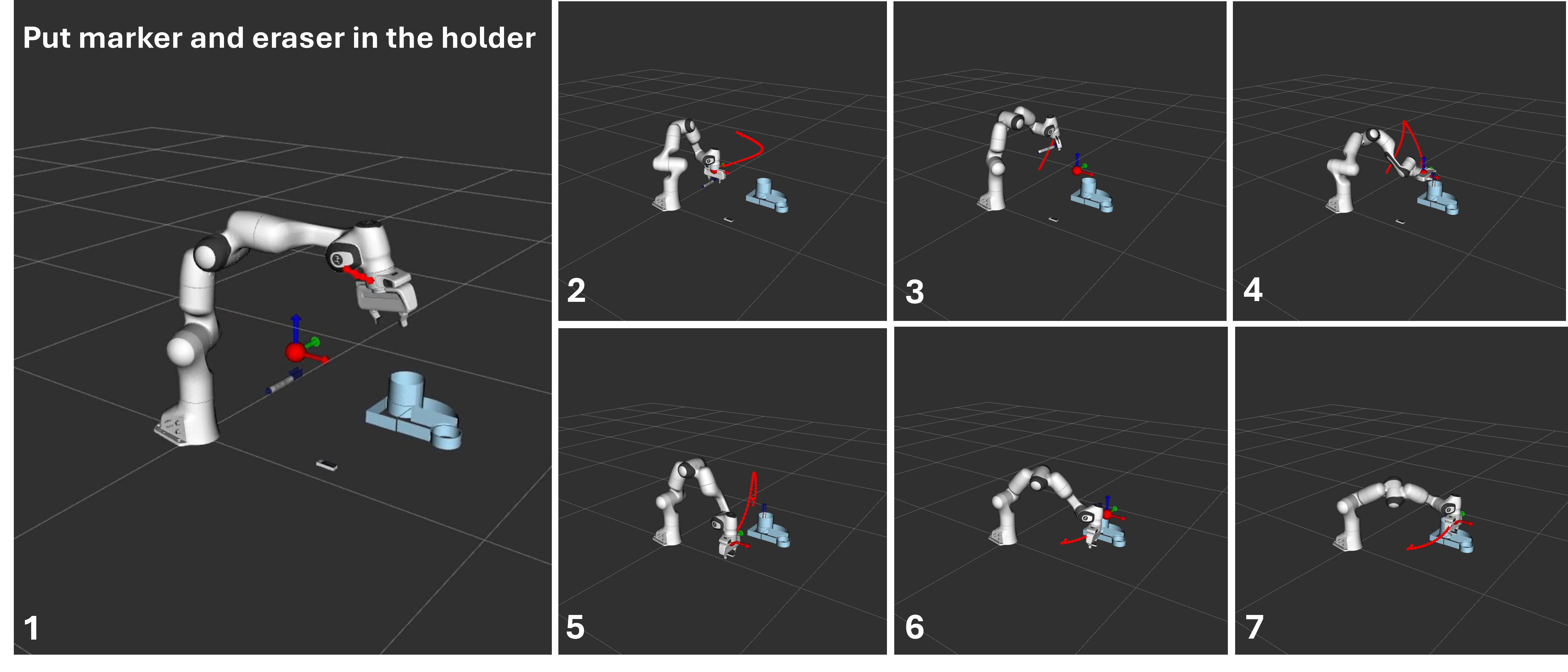}
\caption{Simulation of the Lang2Manip framework performing the task 
``Put the marker and eraser in the holder.'' The sequence illustrates the complete execution pipeline: (1) initial scene with the Panda robot, marker, eraser, and holder; (2) motion toward the marker and grasping; (3) motion towards placement target; (4) placement of the marker into the holder; (5) motion toward the eraser and grasping; (6) motion towards eraser placement target; and (7) final placement of the eraser into the holder. Red curves visualize the Kautham-generated collision-free trajectories for each action.}
\label{fig:snapshot}
\end{figure*}
\subsection{Inverse Kinematics}
Once a desired grasp or placement pose is available, the inverse kinematics (IK) module computes one or more admissible joint configurations,
\(
\bold{q^\star} = \text{IK}(T_{\text{desired}}),
\)
that realize the target end-effector pose. The IK stage is implemented as a modular, plugin-based component, enabling the use of any state-of-the-art IK solver. In our current implementation, we employ the KDL (Kinematics and Dynamics Library) solver, which uses a Newton-Raphson iterative scheme to minimize the pose error between the end-effector and the desired transformation. The solver incorporates joint limits, respects velocity bounds, and leverages redundancy resolution when multiple solutions exist. During each iteration, the Jacobian-based Newton–Raphson update refines the joint configuration until the Cartesian error falls below a specified threshold. If no feasible configuration is found due to kinematic singularities, joint-limit violations, or workspace occlusions, the IK module reports a failure back to the symbolic planning layer, prompting the LLM to reconsider the plan or propose an alternative strategy. This fail-safe loop ensures that the system can adapt when objects are inaccessible, cluttered, or positioned outside the robot's reachable workspace, which improve robustness and overall task success.
\subsection{Planning Query Formulation and Execution in Kautham}

Each validated symbolic action is converted into a motion planning request that is executed through the \texttt{kautham\_ros} interface. Once the start configuration and the goal configuration (computed by the IK module) are available, the system programmatically instantiates a Kautham planning query by specifying the robot model, the set of active obstacles, the planning workspace and the desired motion-planning strategy. Using the \texttt{kautham\_ros} API, the pipeline first selects the appropriate OMPL planner through the \texttt{setPlanner} service, which configures the underlying sampling-based planning algorithm (e.g., RRT, RRTConnect).

After the planner is specified, the symbolic action is mapped to a geometric control query using the \texttt{setQuery} service. This query encodes the initial and final joint configurations. The \texttt{solve}  service is then invoked to compute a feasible trajectory. Internally, Kautham expands the configuration space, performs collision detection against URDF defined environment models, and applies OMPL’s sampling and path construction procedures until a valid trajectory is found.

Once planning query succeeds, the resulting joint path is retrieved through the \texttt{getPath} service. This path is then passed to the trajectory execution component, which interpolates the joint values and forwards them to the robot controller or simulator. Through this ROS-based communication pipeline, Kautham provides a complete cycle of planner configuration, motion-query definition, trajectory computation, and path retrieval, enabling each symbolic action to be grounded in a collision-free geometric trajectory that precisely executes the high-level manipulation plan.

\section{Experiments}
\label{sec:experiments}

\subsection{Experimental Setup}

To evaluate the effectiveness of the Lang2Manip framework, we constructed a simulated manipulation
task using the Franka Emika Panda arm in the Kautham environment. The goal is to 
\textit{put the marker and eraser in the holder}, representing a multi-step task requiring sequential grasping, precise placement, and obstacle-aware trajectory planning. The scene (Fig.~\ref{fig:snapshot}) contains three objects: a marker, an eraser, and a cylindrical holder. All objects are loaded through URDF models specified in the Kautham problem XML file, which also defines their initial poses in the workspace.

The symbolic plan is generated by the LLM from the integrated prompt (task description, system prompt, and textualized Kautham state). Each step of the symbolic plan, pick marker, place marker, pick eraser, place eraser, is transformed into a geometric motion plan using grasp-planning module,  IK solver, and OMPL trajectory planner. The execution sequence is visualized in RViz via \texttt{kautham\_ros}. The figure illustrates all stages of execution, from initial approach to final placement, with red curves showing the collision-free trajectories computed by Kautham.

\subsection{Results}

We evaluate the system according to three commonly used metrics in LLM-TAMP: (i) task success rate, (ii) motion-planning feasibility, and (iii) symbolic-plan correctness.
\textit{Success Rate}
Across 20 trials with randomized initial object poses, Lang2Manip achieved a task completion rate of 85\%, consistent with the range reported in prior LLM-TAMP systems. Most failures were due to unreachable grasps (IK failure) or incorrect pose reasoning by the LLM.
\textit{Motion Feasibility}
motion planning using OMPL within Kautham succeeded in 92\% of the cases, with failures mainly occurring when the LLM provided poses are too close to workspace boundaries or cluttered regions. This confirms that symbolic errors dominate failures rather than the geometric layer.
\textit{Symbolic-Plan Accuracy}
LLM errors, including malformed JSON, missing arguments, or semantically inconsistent action ordering occurred in \textbf{10\%} of the trials. This is comparable to the error rates reported in ProgPrompt \cite{singh2023progprompt} and LLM3.

\section{Discussion}
\label{sec:discussion}

\subsection{Strengths}

The proposed Lang2Manip tool exhibits several strong properties. First, it is \emph{robot-agnostic}: any manipulator with a URDF model can be used without additional engineering. Second, it is \emph{planner-agnostic}: users may switch between geometric, kinodynamic, or physics-based planners in Kautham without modifying code or prompts. Third, symbolic plans from any modern LLM can be executed, as the system imposes no model-specific assumptions. Finally, Kautham’s integration of sampling-based planners, collision models, and multi-view visualization provides a highly transparent environment for understanding LLM-driven manipulation behavior.

Lang2Manip complements existing TAMP and simulation tools. While systems such as MoveIt and PyBullet offer strong collision checking and dynamic simulation, Kautham uniquely provides planning under a unified interface (geometric, kinodynamic, physics-based), as well as multi-robot and workspace-level reasoning. Our integration demonstrates that LLM-based symbolic planners can utilize these advanced capabilities transparently, enabling a broader class of tasks than similar frameworks.
\subsection{Limitations}

Despite its flexibility, the system inherits limitations common to LLM-based task planning. Ambiguity or incompleteness in the LLM output may produce invalid actions or infeasible poses. Simplified grasp heuristics may struggle with irregular shapes or cluttered arrangements. Long-horizon tasks with tightly coupled subgoals sometimes require the LLM to revise or replan symbolically when geometric feasibility checks fail. These challenges motivate deeper feedback loops between symbolic and geometric layers.

\section{Conclusion and Future Work}
\label{sec:conclusion}

This paper presented Lang2Manip, a unified pipeline that bridges LLM-based symbolic planning with the Kautham motion-planning framework to achieve robot-agnostic symbolic-to-geometric planning for manipulation. Through a modular architecture, comprising prompt-driven symbolic plan generation, plugin-based grasp and IK computation, and Kautham’s motion planning back end, the system translates natural-language instructions into executable robot trajectories without robot or planner-specific engineering.
Experimental results validate the robustness and flexibility of the framework and demonstrate performance comparable to recent LLM-TAMP systems.  

Future work will explore hierarchical planning with macro-actions, real-robot deployment using Panda and UR manipulators, integration of learned grasp planners, grounding symbolic actions through multimodal perception (RGB-D, tactile), and closed-loop LLM refinement incorporating uncertainty-aware feedback from the geometric layer.

\balance

\bibliographystyle{IEEEtran}
\bibliography{reference}

\end{document}